\documentclass{article}

 \usepackage{subcaption}
 \usepackage[preprint]{neurips_2025}

\usepackage[utf8]{inputenc} 
\usepackage[T1]{fontenc}    
\usepackage{hyperref}       
\usepackage{url}            
\usepackage{booktabs}       
\usepackage{amsfonts}       
\usepackage{nicefrac}       
\usepackage{microtype}      
\usepackage{xcolor}         
\usepackage{enumitem}
\usepackage{graphicx}
\usepackage{makecell}
\usepackage{multirow}
\usepackage{enumitem}
\usepackage{wrapfig}
\usepackage{tcolorbox}
\tcbuselibrary{listingsutf8}

\title{Do Cognitively Interpretable Reasoning Traces Improve LLM Performance?}

\author{%
  Siddhant Bhambri, \\
  SCAI,
  Arizona State University,\\
  Tempe, US \\
  \texttt{sbhambr1@asu.edu}  
  \And
  Upasana Biswas, \\
  SCAI,
  Arizona State University,\\
  Tempe, US\\
  \texttt{ubiswas2@asu.edu} 
  \And
  Subbarao Kambhampati,\\
  SCAI,
  Arizona State University,\\
  Tempe, US\\
  \texttt{rao@asu.edu} 
}

\begin{document}

\maketitle

\begin{abstract}
Recent progress in reasoning-oriented Large Language Models (LLMs) has been driven by introducing Chain-of-Thought (CoT) traces, where models generate intermediate reasoning traces before producing an answer. These traces, as in DeepSeek R1, are not only used to guide inference but also serve as supervision signals for distillation into smaller models. A common but often implicit assumption is that CoT traces should be semantically meaningful and interpretable to the end user. While recent research questions the need for semantic nature of these traces, in this paper, we ask: ``\textit{Must CoT reasoning traces be interpretable to enhance LLM task performance?}" We investigate this question in the Open Book Question-Answering domain by supervised fine-tuning LLaMA and Qwen models on four types of reasoning traces: (1) DeepSeek R1 traces, (2) LLM-generated summaries of R1 traces, (3) LLM-generated post-hoc explanations of R1 traces, and (4) algorithmically generated verifiably correct traces. To quantify the trade-off between interpretability and performance, we further conduct a human-subject study with 100 participants rating the interpretability of each trace type. Our results reveal a striking mismatch: while fine-tuning on R1 traces yields the strongest performance, participants judged these traces to be the least interpretable. These findings suggest that it is useful to decouple intermediate tokens from end user interpretability.
\end{abstract}
\section{Introduction}
\label{sec:introduction}
Reasoning with intermediate Chain-of-Thought (CoT)-style traces has become one of the defining strategies for improving the performance of Large Language Models over a diverse range of problems, as popularized by DeepSeek R1 \cite{guo2025deepseek}. While models such as DeepSeek R1 often generate excessively verbose responses \cite{kambhampati2025reasoning}, the R1-generated reasoning traces have been utilized as a learning signal for Supervised Fine-Tuning (SFT) smaller models to boost their performance \cite{magister2022teaching,Shridhar2022DistillingRC,tian2025beyond}. 

A common but often implicit assumption behind these CoT traces is that they should be semantically meaningful and interpretable to humans. Although training with these traces is done primarily to improve LLM performance on a given task, these traces need not be semantically correct or interpretable to optimize this objective. Moreover, since these reasoning traces are exposed to the end user, they can possibly exacerbate issues like user distrust, misinformation, errors, and perpetuated biases \cite{guidotti2018survey}. This distinction has also been brought to light by the recent GPT-OSS models that generate a CoT trace, a summary, and the final answer where the summary is shown for the humans and not the CoT trace \cite{gpt-oss}. There has been recent work that has challenged the first assumption behind these traces to be semantically meaningful by showing that both transformers and pre-trained LLMs can perform better when trained (or fine-tuned) with semantically incorrect traces paired with correct final solutions \cite{bhambri2025interpretable,stechly2025beyond}. In this work, we aim to specifically want to answer - ``\textit{Must CoT reasoning traces be interpretable to enhance LLM task performance?}".

We specifically look at the Open Book Question Answering domain and utilize the CoTemp QA benchmark \cite{su2024living} which consists of problems comprising a set of facts that can be utilized to answer the respective question. We conduct Supervised Fine-Tuning (SFT) experiments on Llama-3.2-1B-Instruct, Llama-3.1-8B, and Qwen3-1.7B, and Qwen3-8B chat models using four different variations of reasoning traces paired with correct final solutions. We consider (1) DeepSeek R1 traces, (2) LLM (GPT-4o-mini)-generated summaries of R1 traces, (3) LLM (GPT-4o-mini)-generated post-hoc explanations of R1 traces, and (4) algorithmically generated verifiably correct traces (as introduced in \cite{bhambri2025interpretable} for Open Book QA benchmarks). We then compare the final solution accuracy across all fine-tuned models.

To objectively compare the interpretability across each of these trace types, we further conduct a human-subject study with 125 participants. Five different sets of 25 participants were hired on Prolific and asked to rate each of the reasoning trace types on a Likert Scale measuring interpretability via attributes such as reasoning trace predictability, comprehensibility, and faithfulness~\cite{user-2, user-1}.  
While fine-tuning on R1 traces shows the strongest performance on three out of the four LLMs, our striking results reveal that users find R1 traces to be the least interpretable across all tested attributes when compared with the other trace types. R1 traces scored the lowest among all variations of reasoning traces, averaging 3.396 among all interpretability attributes. These findings highlight that cognitive interpretability of reasoning traces can in fact be an albatross from the perspective of LLM's task performance.

\section{Related Work}
\label{sec:related_work}
Large Language Models have significantly benefited from training with CoT  traces coupled with final solutions for a variety of problems. There have been studies that have looked at and argued for making these CoT traces more interpretable, aka improve their faithfulness for the end user, as they are believed to serve as the LLM's explanations to generate the final solution \cite{arcuschin2025chain,tanneru2024hardness,li2024towards,tutek2025measuring,paul2024making,lyu2023faithful,lanham2023measuring,yeo2024interpretable}. On the other hand, there has also been work showcasing why these traces are not explainable to the end user \cite{barez2025chain}. Both sides of this argument stem from the assumption that these traces are indeed meant to be useful for the end user and not just for the LLM to improve its final solution performance over a certain task. We specifically challenge this assumption in an effort to show the disconnect between the use of CoT traces for the LLM (as a training signal in SFT) and the use of CoT traces for the end user (as an interpretable reason behind the model's final solution). Interpretability has been studied along multiple attributes, including predictability, comprehensibility, interpretability, and faithfulness \cite{user-1,user-2}. These dimensions have been used to evaluate post hoc XAI methods and to develop a rigorous framework for interpretable machine learning.

\section{Measuring LLM Performance via SFT w/ different Reasoning Traces}
\label{sec:sft_experiments_results}

\begin{wrapfigure}{r}{0.5\columnwidth}
    \centering
    \includegraphics[width=0.48\columnwidth]{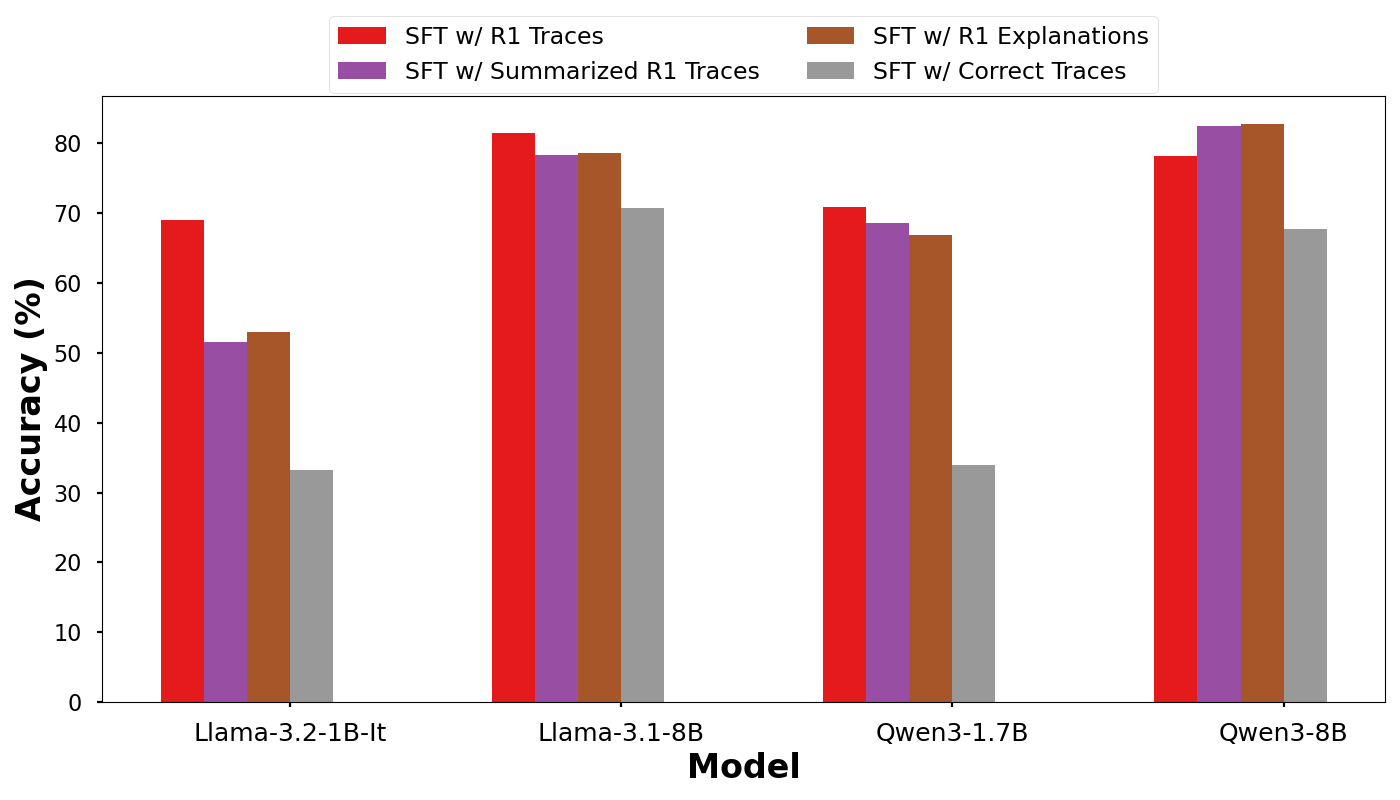}
    \caption{Final solution performance on CoTemp QA test dataset after SFT with different trace types on Llama and Qwen models.}
    \label{fig:trace_sft_comparison}
\end{wrapfigure}
\textbf{Dataset and Metrics:} CoTemp QA \cite{su2024living} consists of English co-temporal questions which involve identifying the type of temporal relation posed in the problem, followed by inferring which fact in the given passage of text satisfies the temporal relation with the question. For all our SFT experiments, we look at the final solution accuracy across all fine-tuned models using the four trace types.

\textbf{Reasoning Trace generation:} We first consider (1) DeepSeek R1 traces where we prompt the R1 model on the CoTemp QA training dataset and collect the model responses for our SFT experiments where it got the correct final answer. Utilizing this filtered training dataset, we prompt GPT-4o-mini to generate both (2) summaries and (3) post-hoc explanations of these R1 traces. Since R1 traces can often be verbose, we posit that their summary as well as a post-hoc explanation can likely be more interpretable to the end user. As a control study, we utilize the generated SFT trace datasets from \cite{bhambri2025interpretable} which consist of the (4) semantically correct verifiable reasoning traces. These traces have been algorithmically generated by extracting the relevant fact/s from the provided passage.


\textbf{Results:} 
We highlight the key SFT results in Figure \ref{fig:trace_sft_comparison}. A common observation seen across three out of the four models (except in Qwen3-8B) is that SFT with R1 traces leads to the highest final solution accuracy over SFT with any other trace type with the largest performance boost seen for Llama-3.2-1B-Instruct model. Furthermore, among all the four models, we note that SFT with the algorithmically-generated semantically correct traces and SFT with the adversarially-generated incorrect traces perform the worst also in comparison to SFT with summaries and explanations of R1 traces. Keeping these results in consideration, we conduct a user study to test if the R1 traces that led to the best performing SFT models are interpretable to humans.

\section{Measuring Trace Interpretability via Human-Subject Studies}
\label{sec:user_study_expt_results}
We conducted four separate user studies to evaluate the interpretability of the four types of reasoning traces. In each study, participants were shown only one type of trace: (1) DeepSeek R1 traces, (2) summarized R1 traces, (3) post-hoc explanations of R1 traces, or (4) verifiably correct reasoning races \cite{bhambri2025interpretable}.  Our hypotheses are tested in a between-subjects design, comparing participant responses across these four groups. The specific hypotheses we test are:
\begin{itemize}[noitemsep, topsep=0pt, label=-]
\item \textbf{H1:} Reasoning traces that improve task accuracy will not lead to higher interpretability for the user.
\item \textbf{H2:} Reasoning traces that improve task accuracy will be associated with higher cognitive workload for the user, as measured by increased mental demand, effort, and frustration.
\end{itemize}
\paragraph{Experimental Setup:}For each trace type, we recruited 25 users (100 in total) on Prolific. Each participant viewed five Q/A examples (fixed across all studies), consisting of the input question, the predicted answer, and the reasoning trace. After each example, participants rated the reasoning trace on a 5-point Likert scale on the following properties as suggested by ~\cite{user-2, user-1}: predictability, comprehensibility, interpretability, and faithfulness to context (alignment with given facts). To capture the cognitive workload involved in processing and evaluating the traces, we used the NASA–TLX assessment \cite{hart2006nasa}, focusing on the dimensions of mental demand, effort, and frustration.  \\
\textbf{Main Findings:}
From Table~\ref{user-study}, we observe that participants rated algorithmically generated correct reasoning traces from \cite{bhambri2025interpretable} as the most interpretable across all dimensions—predictability, comprehensibility, interpretability, and faithfulness—consistently scoring higher medians than all other trace types. In contrast, R1 traces received the lowest interpretability ratings across every dimension. Summarized R1 traces and R1-trace explanations received intermediate ratings, indicating that compact or post-hoc representations improve human comprehension compared to raw R1 traces. In terms of cognitive workload, R1 traces imposed higher mental demand, effort, and frustration compared to other kinds of traces. Correct traces were associated with relatively lower cognitive workload, indicating that users found them easiest to follow and comprehend. \\
\textbf{Statistical Analysis: }
We conducted pairwise Mann–Whitney U tests \cite{mcknight2010mann} at a significance level of $\alpha=0.05$ with Bonferroni correction applied for multiple comparisons.
Null hypotheses derived from our study hypotheses were defined as follows:
NH1 (Interpretability): There is no difference in interpretability ratings between R1 traces and algorithmically-generated correct reasoning traces.
NH2 (Cognitive Workload): There is no difference in cognitive workload ratings between R1 traces and algorithmically-generated correct reasoning traces.\\
There was a significant difference in interpretability measured between these two trace types, across all measured dimensions (predictability: U = 176.5 ,$p =.00022 < 0.05$; comprehensibility: U =  175, $p =.00019 < 0.05$; interpretability: U = 161, $p =  .00014 < 0.05$; faithfulness: U = 178.5, $p =  .00015 < 0.05$). Further analysis also shows that there was a significant difference between cognitive workload of users between the two trace types, across all measured dimensions (mental demand: U = 194, $p = .00036 < 0.05$; effort: U =  176, $p =.00013 < 0.05$; frustration: U = 176.5, $p =  .01287 < 0.05$). Thus, we can reject NH1 and NH2, validating 

\begin{table}[ht!]
\centering
\renewcommand{\arraystretch}{1.0}
\setlength{\tabcolsep}{2pt} 

\begin{tabular}{l | p{0.32\columnwidth} | c c c c} 
\toprule
\textbf{Dimension} & \makecell{\textbf{Question}} & 
\makecell{\textbf{R1} \\ \textbf{Traces}} & 
\makecell{\textbf{Summarized} \\ \textbf{R1 Traces}} & 
\makecell{\textbf{R1} \\ \textbf{Explanations}} & 
\makecell{\textbf{Correct} \\ \textbf{Traces}} \\
\midrule
Predictability & \footnotesize I could anticipate the next steps or conclusions based on earlier parts of the reasoning ↑ & \textbf{3.48} & 4.45 & 4.29 & 4.82 \\
Comprehensibility & \footnotesize I understood the reasoning followed by the model ↑ & \textbf{3.46} & 4.55 & 4.27 & 4.56 \\
 & \footnotesize I could follow each step in the reasoning without confusion ↑ & \textbf{3.46} & 4.54 & 4.28 & 4.84 \\
Interpretability & \footnotesize The reasoning helped me understand why the model acted or concluded the way it did ↑ & \textbf{3.31} & 4.53 & 4.29 & 4.86 \\
Faithfulness & \footnotesize There were no major gaps or missing reasoning steps in the reasoning ↑ & \textbf{3.33} & 4.54 & 4.26 & 4.72 \\
 & \footnotesize The reasoning is consistent with the facts or evidence provided in the context ↑ & \textbf{3.34} & 4.24 & 4.29 & 4.84 \\
\midrule
Mental Demand & \footnotesize How mentally demanding was the task? ↓ & \textbf{4.65} & 2.87 & 2.92 & 2.31 \\
Effort & \footnotesize How hard did you have to work to accomplish your level of performance? ↓ & \textbf{4.54} & 2.39 & 2.17 & 2.86 \\
Frustration & \footnotesize How frustrated, stressed, and annoyed were you? ↓ & \textbf{4.58} & 2.04 & 2.42 & 2.42 \\
\bottomrule
\end{tabular}

\vspace{2mm} 
\caption{Median participant ratings of reasoning traces across dimensions of interpretability and cognitive workload. Arrows indicate the desired direction of scores: ↑ higher ratings are better for interpretability measures, ↓ lower ratings are better for cognitive workload measures.}
\label{user-study}
\end{table}
\section{Discussion}
\label{sec:discussion}
Our findings reveal a major disconnect between the utility of reasoning traces for improving LLM performance and their cognitive interpretability for humans. SFT with R1 traces led to higher accuracy, yet these traces were rated the lowest across all dimensions of interpretability—predictability, comprehensibility, and faithfulness—and were associated with the highest mental demand, effort, and frustration.  Summaries and post-hoc explanations of R1 traces further validate this point: although they yielded lower accuracy than R1 traces, they are easier for users to predict, understand, and perceive as faithful. By contrast, algorithmically generated correct traces were judged as most interpretable and least mentally demanding, but yielded the weakest improvements in model accuracy. These results clearly highlight that verbose traces like R1 provide rich training signals for models, but are poorly aligned with the interpretability expectations of the end user. Furthermore, the reasoning traces which benefit the LLMs the most need not have semantic structure, underscoring a fundamental disconnect between what serves as a good training signal and what supports human understanding.
\section{Conclusion}
\label{sec:conclusion}

In this work, we studied the relationship between the use reasoning traces for improving model performance and their interpretability for end users. Through SFT experiments on four LLMs, we observed that R1 traces yielded the highest accuracy on the CoTemp QA benchmark over other trace types. In contrast, our human-subject study revealed that these same R1 traces were rated lowest across all interpretability dimensions and imposed the greatest cognitive workload. These findings demonstrate that reasoning traces which help models do not necessarily carry semantics that humans find interpretable. This decoupling broadly has two key takeaways for future works - (1) CoT-style traces should only be utilized for optimizing model performance and not end-user interpretability, and (2) independent efforts should be carried out for generating explanations behind the model's answer tailored for the end-user.

\section{Acknowledgments}
This research is supported primarily by ONR grant N0001423-1-2409. This research is also supported by DARPA grant HR00112520016, and gifts from Qualcomm, J.P. Morgan and Amazon.
\bibliography{latex/custom}
\bibliographystyle{plain}


\appendix


\newpage
\section{Additional Experiment Details}

\subsection{Dataset}
\label{subsec:appendix_dataset}

\paragraph{CoTemp QA:} The dataset is categorized into four temporal relation types, namely - `equal', `overlap', `during' and `mix', and requires around one or two facts for answering the question. For our experiments, we utilize 3,798 train and 950 test samples to construct the SFT datasets. The train/test splits for each category are shown in Table \ref{tab:cotemp_data}.

\begin{table}[ht]
\centering
\caption{Train and Test data distribution for CoTemp QA dataset used in our SFT experiments.}
\label{tab:cotemp_data}
\begin{tabular}{@{}c|c@{}}
\textbf{Category} & \textbf{Train/Test Samples} \\ \midrule
equal             & 349 / 87                    \\
overlap           & 522 / 131                   \\
during            & 2477 / 619                  \\
mix               & 450 / 113                   \\ \bottomrule
\end{tabular}%
\end{table}

\subsection{Implementation Details and Hyper-parameters}
\label{subsec:appendix_expts_implementation}
Models were fine-tuned using the Hugging Face library \cite{wolf-etal-2020-transformers} on a single 80GB NVIDIA Tesla A100 GPU for 3 epochs (effective batch size 16, max sequence length 1024). We employed PEFT QLoRA \cite{dettmers2023qlora} (rank 16, alpha 32) with a learning rate of 2e-4 (8-bit AdamW, cosine scheduler, 0.1 warm-up). Prompt experiments utilized vLLM \cite{kwon2023efficient}. We will release the code and datasets used for our experiments on acceptance.

\subsection{Prompts}
\label{sec:appendix_prompts}

\begin{tcolorbox}[
    colback=gray!5!white, 
    colframe=black, 
    title=R1 Trace Summarization Prompt, 
    fonttitle=\bfseries, 
    sharp corners, 
    boxrule=0.8pt,
    left=4pt, right=4pt, top=4pt, bottom=4pt
]
Summarize the following trace in a very concise and clear manner, highlighting key events and outcomes in less than 100 words: 

...

\textbf{\{R1 trace\}}

...

\textbf{Summary:}
\end{tcolorbox}

\begin{tcolorbox}[
    colback=gray!5!white, 
    colframe=black, 
    title=R1 Trace Explanation Prompt, 
    fonttitle=\bfseries, 
    sharp corners, 
    boxrule=0.8pt,
    left=4pt, right=4pt, top=4pt, bottom=4pt
]
\textbf{\{Problem\}}

...

\textbf{\{R1 trace\}}

...

\textbf{\{R1 answer\}}

You have answered the question correctly. Please provide a detailed explanation of the reasoning behind your answer. The explanation should be clear, concise, and easy to understand.

...

\textbf{Explanation:}
\end{tcolorbox}
\section{User Study}
To evaluate the interpretability of reasoning traces generated by reasoning models, we conducted a set of structured user studies. Each participant was given a compensation of $12\$/hr$. The IRB protocol details will be released on acceptance. Each study followed the same sequence of steps, designed to ensure consistency across participants for each study. Below we outline the main components of the study design. 
\subsection{Human Participant Demographics}
We conducted four user studies with participants recruited through Prolific (all located in the United States). In general, the participant populations in all four studies were demographically similar, with no major differences in the age or education distribution, suggesting that the results in the studies are comparable and not driven by differences in the composition of the participants.
\paragraph{Education:}  
Participants spanned a range of educational backgrounds. Across all studies, the majority held an \emph{Undergraduate Degree} (roughly 45--55\% in each study), followed by \emph{Master's Degrees} (20--30\%), and a smaller proportion with \emph{PhDs or equivalent doctoral-level degrees} (10--15\%). A minority of participants reported \emph{High School}, \emph{Associate's Degree}, or \emph{Some College} as their highest level of education ($<$10\% each). These proportions were consistent across the four studies. 
\paragraph{Age:} The participants were distributed over a wide age range, with the largest groups being \emph{35--50 years old} (approximately 35--40\%) and \emph{51+ years old} (30--35\%). Younger age groups were represented to a lesser extent: \emph{26--34 years old} (20--25\%) and \emph{18--25 years old} (5--10\%). Again, these proportions were stable across studies. 
\subsection{Consent and Statement}
Each participant began the study by reviewing and agreeing to a consent statement. The statement explained the goals of the study, what participants would be asked to do, and how their data would be handled.
\begin{figure}[h]
    \centering
    \begin{minipage}{0.48\linewidth}
        \centering
        \fbox{\includegraphics[width=\linewidth]{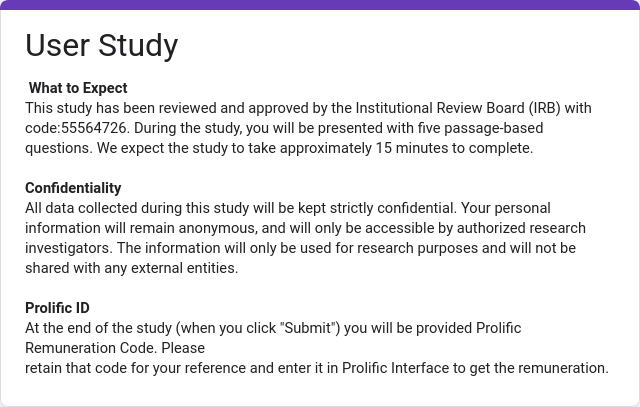}}
        \caption{Consent statement shown to participants before starting the study.}
    \end{minipage}%
    \hfill
    \begin{minipage}{0.48\linewidth}
        \centering
        \fbox{\includegraphics[width=\linewidth]{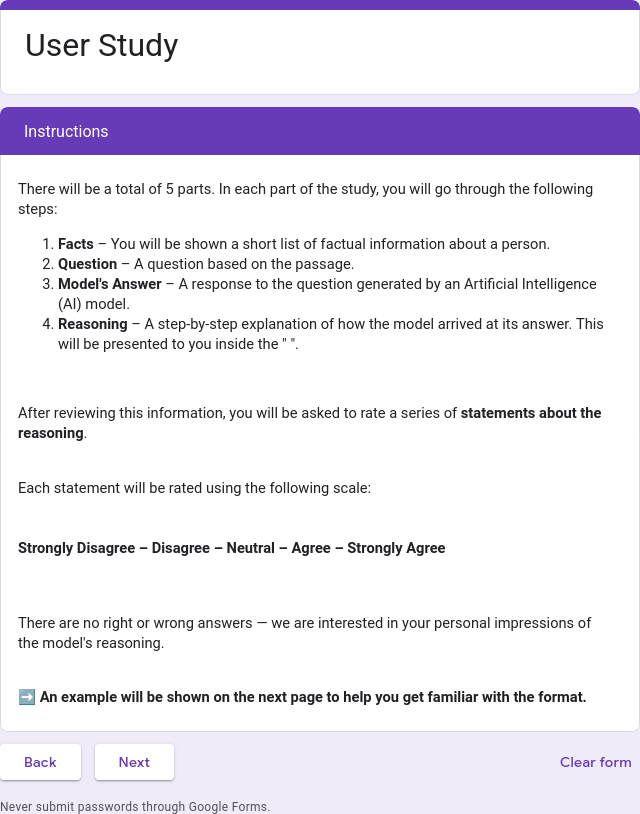}}
        \caption{Instructions shown to participants before starting the study.}
    \end{minipage}
    \caption{Consent statement (left) and instructions (right) shown to participants.}
\end{figure}

\subsection{Instructions}
Participants were provided with detailed instructions describing the study structure. Each of the five parts of the study followed the same format:
\begin{enumerate}
    \item \textbf{Facts:} A short list of factual statements about a person.  
    \item \textbf{Question:} A query based on the passage.  
    \item \textbf{Model's Answer:} The response generated by the AI model.  
    \item \textbf{Reasoning:} A step-by-step explanation of how the model arrived at its answer.  
\end{enumerate}
After reviewing this information, participants rated statements about the reasoning on a 5-point Likert scale (Strongly Disagree–Strongly Agree).  
\begin{figure}[h]
    \centering
    \begin{minipage}{0.48\linewidth}
        \centering
        \fbox{\includegraphics[width=\linewidth]{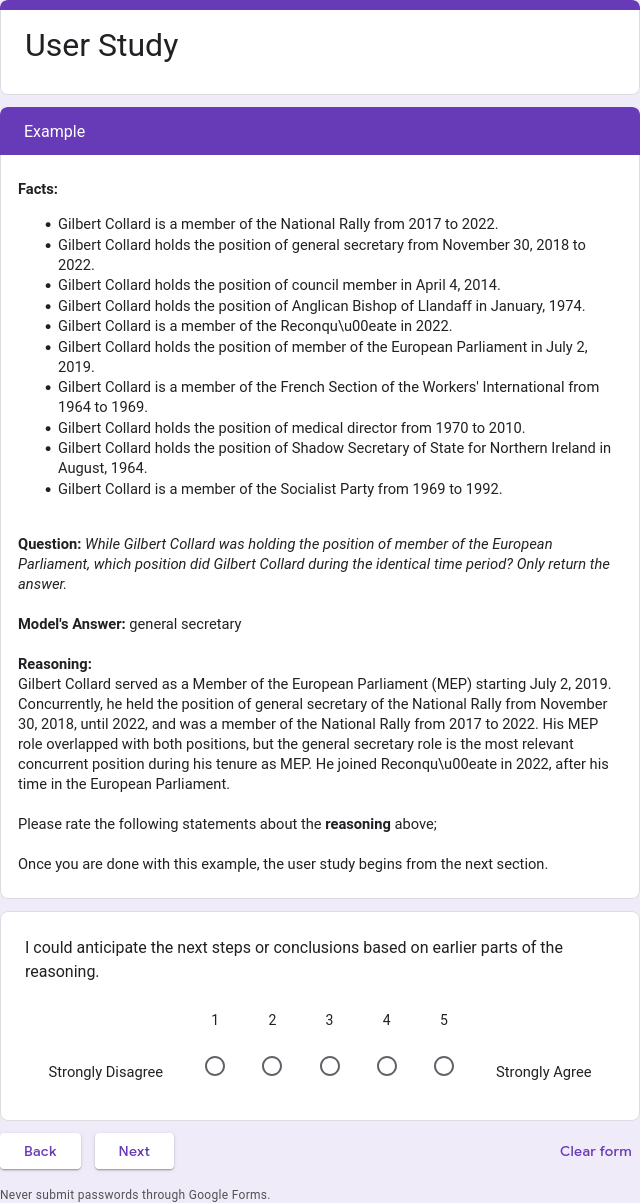}}
        \caption{Example shown to participants.}
    \end{minipage}%
    \hfill
    \begin{minipage}{0.48\linewidth}
        \centering
        \fbox{\includegraphics[width=\linewidth]{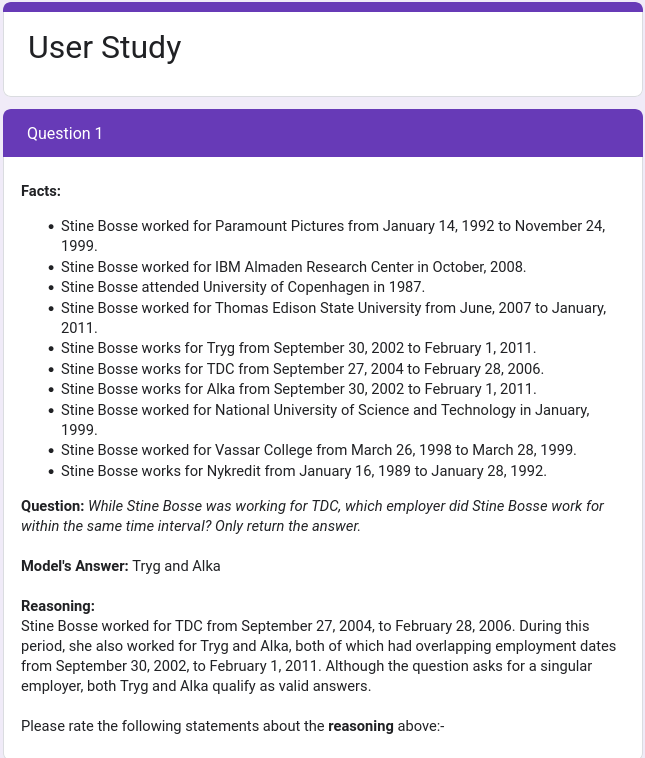}}
        \caption{Task shown to participants.}
    \end{minipage}
    \caption{Example (left) and task (right) shown to participants.}
\end{figure}
\subsection{Q/A Task}
Before beginning the main task, participants reviewed an example question and answer with reasoning (see Fig.~X). Participants then completed five Q/A tasks of the same form as the example. Each task included a passage, model answer, reasoning trace, and associated questionnaire.  
\begin{figure}[h]
    \centering
    \begin{minipage}{0.48\linewidth}
        \centering
        \fbox{\includegraphics[width=\linewidth]{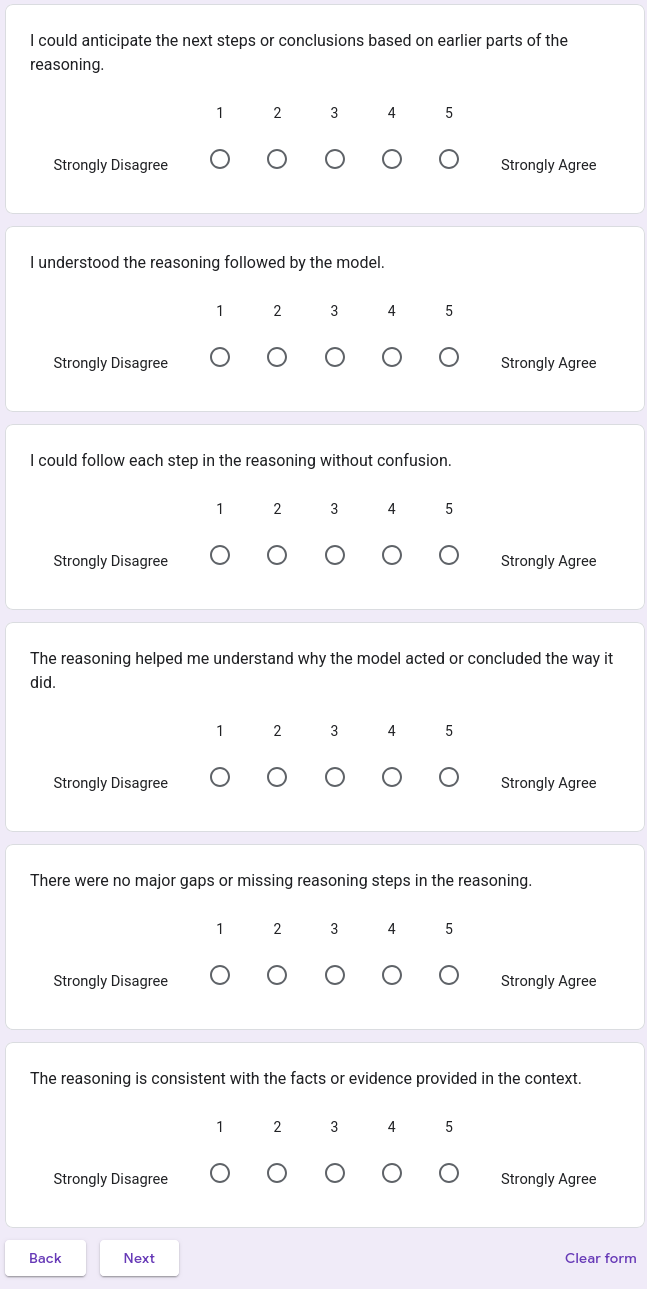}}
    \end{minipage}%
    \hfill
    \begin{minipage}{0.48\linewidth}
        \centering
        \fbox{\includegraphics[width=\linewidth]{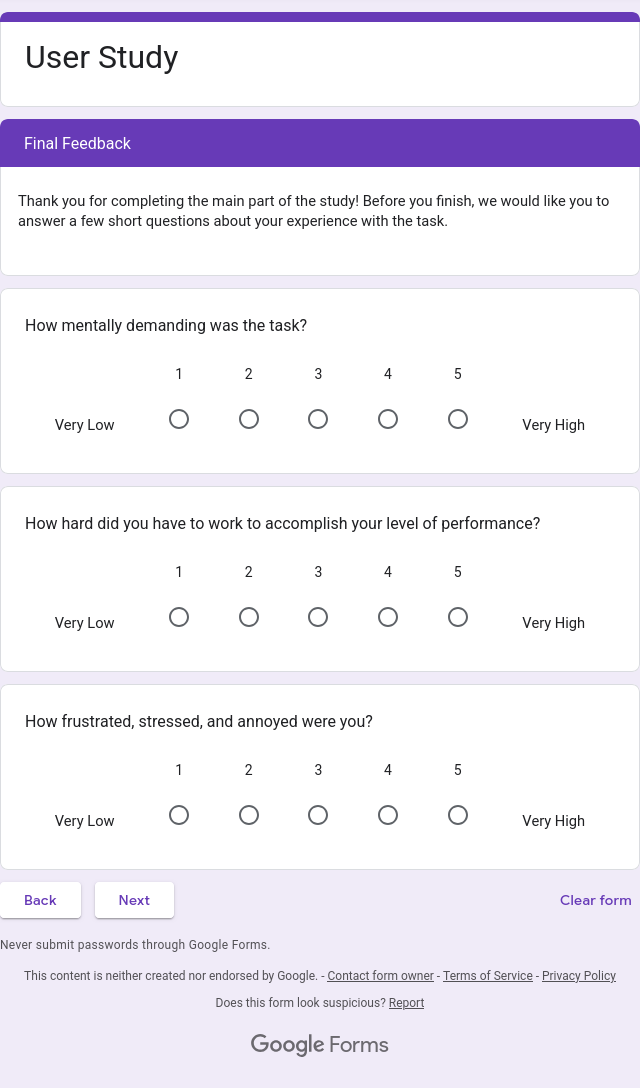}}
    \end{minipage}
\end{figure}
\subsection{Questionnaire}
After each reasoning trace, participants filled out a brief questionnaire assessing dimensions of interpretability (see Fig.~Y).  
At the end of the study, participants completed a feedback survey, with NASA-TLX questions to measure perceived workload.  
\newpage
\subsection{Statistical Analysis Results}
\begin{table*}[ht!]
\centering
\small
\renewcommand{\arraystretch}{1.5} 
\setlength{\tabcolsep}{8pt} 

\begin{tabular}{l|ccc|ccc|ccc}
\toprule
Dimension & 
\multicolumn{3}{c|}{R1 vs Correct} & 
\multicolumn{3}{c|}{R1 vs Summarized R1} & 
\multicolumn{3}{c}{R1 vs Explanations} \\
 & $U$ & $p$ & Sig. & $U$ & $p$ & Sig. & $U$ & $p$ & Sig. \\
\midrule
\multicolumn{10}{c}{\textbf{H1: Interpretability}} \\
Predictability      & 176.5 & .00022 & Yes & 177.0 & .00036 & Yes & 126.5 & .0004 & Yes \\
Comprehensibility   & 175   & .00019 & Yes & 102.2 & $<.00001$ & Yes & 126 & .0006 & Yes \\
Interpretability    & 161   & .00014 & Yes & 74.5 & $<.00001$ & Yes & 187 & $<.00001$ & Yes \\
Faithfulness        & 178.5 & .00015 & Yes & 73.5 & $<.00001$ & Yes & 115.5 & $<.00001$ & Yes \\
\midrule
\multicolumn{10}{c}{\textbf{H2: Cognitive Workload}} \\
Mental Demand       & 194   & .00036 & Yes & 237.5 & 0.055 & No & 264 & .21 & No \\
Effort              & 176   & .00013 & Yes & 169.5 & .0016 & Yes & 104 & $<.00001$ & Yes \\
Frustration         & 102.5 & .01287 & Yes & 164 & 0.0013 & Yes & 158 &  .00056 & Yes \\
\bottomrule
\end{tabular}
\caption{Pairwise Mann--Whitney U test results across different trace types. Significance is determined at $\alpha=0.05$ with Bonferroni correction.}
\label{tab:mann-whitney-all-traces}
\end{table*}

\section{Limitations}
In this work, we analyze the correlation between the final solution performance of LLMs when fine-tuned with different types of CoT-style traces and the interpretability of these traces for end users. Due to computational limitations, we restrict our scope on experiments with models up to 8 billion parameters. We also restrict our study on simple Open Book QA problems that can be easily understood and answered by lay users. Future works can scale our study to analyze the impact on final performance by fine-tuning larger parameter LLMs, and hire domain experts for conducting user studies on scientific benchmarks such as math or coding problems.


\newpage
\section*{NeurIPS Paper Checklist}

\begin{enumerate}

\item {\bf Claims}
    \item[] Question: Do the main claims made in the abstract and introduction accurately reflect the paper's contributions and scope?
    \item[] Answer: \answerYes{} 
    \item[] Justification: The claims made in the abstract and introduction are clearly defined in the later parts of the paper substantiated with the respective experiment results and findings.

\item {\bf Limitations}
    \item[] Question: Does the paper discuss the limitations of the work performed by the authors?
    \item[] Answer: \answerYes{} 
    \item[] Justification: The limitations of the work have been included in the Appendix of the paper.

\item {\bf Theory assumptions and proofs}
    \item[] Question: For each theoretical result, does the paper provide the full set of assumptions and a complete (and correct) proof?
    \item[] Answer: \answerNA{} 
    \item[] Justification: N/A

    \item {\bf Experimental result reproducibility}
    \item[] Question: Does the paper fully disclose all the information needed to reproduce the main experimental results of the paper to the extent that it affects the main claims and/or conclusions of the paper (regardless of whether the code and data are provided or not)?
    \item[] Answer: \answerYes{} 
    \item[] Justification: All necessary experiment details have been included in Section 3 and 4 of the main paper along with the Appendix.

\item {\bf Open access to data and code}
    \item[] Question: Does the paper provide open access to the data and code, with sufficient instructions to faithfully reproduce the main experimental results, as described in supplemental material?
    \item[] Answer: \answerYes{} 
    \item[] Justification: Both code and data will be released, current submission does not allow attaching data and code in supplementary material.

\item {\bf Experimental setting/details}
    \item[] Question: Does the paper specify all the training and test details (e.g., data splits, hyperparameters, how they were chosen, type of optimizer, etc.) necessary to understand the results?
    \item[] Answer: \answerYes{} 
    \item[] Justification: All necessary experiment details have been provided in the Appendix.

\item {\bf Experiment statistical significance}
    \item[] Question: Does the paper report error bars suitably and correctly defined or other appropriate information about the statistical significance of the experiments?
    \item[] Answer: \answerYes{} 
    \item[] Justification: Statistical significance tests have been performed for the user study results.

\item {\bf Experiments compute resources}
    \item[] Question: For each experiment, does the paper provide sufficient information on the computer resources (type of compute workers, memory, time of execution) needed to reproduce the experiments?
    \item[] Answer: \answerYes{} 
    \item[] Justification: Experiment compute resource details have been provided in the Appendix.
    
\item {\bf Code of ethics}
    \item[] Question: Does the research conducted in the paper conform, in every respect, with the NeurIPS Code of Ethics \url{https://neurips.cc/public/EthicsGuidelines}?
    \item[] Answer: \answerYes{} 
    \item[] Justification: NeurIPS Code of Ethics has been followed to conduct this research.

\item {\bf Broader impacts}
    \item[] Question: Does the paper discuss both potential positive societal impacts and negative societal impacts of the work performed?
    \item[] Answer: \answerYes{} 
    \item[] Justification: Broader impacts of this work have been discussed in the Introduction and Conclusion sections.
    
\item {\bf Safeguards}
    \item[] Question: Does the paper describe safeguards that have been put in place for responsible release of data or models that have a high risk for misuse (e.g., pretrained language models, image generators, or scraped datasets)?
    \item[] Answer: \answerNA{} 
    \item[] Justification: N/A

\item {\bf Licenses for existing assets}
    \item[] Question: Are the creators or original owners of assets (e.g., code, data, models), used in the paper, properly credited and are the license and terms of use explicitly mentioned and properly respected?
    \item[] Answer: \answerYes{}{} 
    \item[] Justification: Citations and references have been provided everywhere where necessary.

\item {\bf New assets}
    \item[] Question: Are new assets introduced in the paper well documented and is the documentation provided alongside the assets?
    \item[] Answer: \answerYes{} 
    \item[] Justification: The code that will be released with the paper will be well documented for easy reproducibility.

\item {\bf Crowdsourcing and research with human subjects}
    \item[] Question: For crowdsourcing experiments and research with human subjects, does the paper include the full text of instructions given to participants and screenshots, if applicable, as well as details about compensation (if any)? 
    \item[] Answer: \answerYes{} 
    \item[] Justification: All human subject study details and compensation have been discussed in the Appendix.

\item {\bf Institutional review board (IRB) approvals or equivalent for research with human subjects}
    \item[] Question: Does the paper describe potential risks incurred by study participants, whether such risks were disclosed to the subjects, and whether Institutional Review Board (IRB) approvals (or an equivalent approval/review based on the requirements of your country or institution) were obtained?
    \item[] Answer: \answerYes{} 
    \item[] Justification: There were no risks posed to the human subject study participants, and the study was approved under the respective IRB protocol.

\item {\bf Declaration of LLM usage}
    \item[] Question: Does the paper describe the usage of LLMs if it is an important, original, or non-standard component of the core methods in this research? Note that if the LLM is used only for writing, editing, or formatting purposes and does not impact the core methodology, scientific rigorousness, or originality of the research, declaration is not required.
    \item[] Answer: \answerYes{} 
    \item[] Justification: All LLM-usage related details have been clearly provided in the paper where necessary.

\end{enumerate}
\end{document}